# A kernel for time series based on global alignments

Marco Cuturi[1]*, Jean-Philippe Vert[2],
Øystein Birkenes[3], Tomoko Matsui[1].

[1]Institute of Statistical Mathematics
Tokyo, Japan

[2]Ecole des Mines de Paris,
ParisTech
Paris, France

[3]Norwegian University of
Science and Technology
Trondheim, Norway

October 2, 2006

**Abstract**

We propose in this paper a new family of kernels to handle times series, notably speech data, within the framework of kernel methods which includes popular algorithms such as the Support Vector Machine. These kernels elaborate on the well known Dynamic Time Warping (DTW) family of distances by considering the same set of elementary operations, namely substitutions and repetitions of tokens, to map a sequence onto another. Associating to each of these operations a given score, DTW algorithms use dynamic programming techniques to compute an optimal sequence of operations with high overall score. In this paper we consider instead the score spanned by all possible alignments, take a smoothed version of their maximum and derive a kernel out of this formulation. We prove that this kernel is positive definite under favorable conditions and show how it can be tuned effectively for practical applications as we report encouraging results on a speech recognition task.

## 1 Introduction

Defining adequate kernels to handle properly structured objects, and notably time series, remains a key challenge for practitioners interested in the application of kernel methods to real-life data-sets. While practitioners willing to use kernel machines are tempted to apply standard vector kernels on time series, such as the popular Gaussian and polynomial kernels implemented in most toolboxes, they are faced with two issues: first, the time series considered in their databases might be of variable length, and second, standard kernels for vectors cannot capture by construction the local dependencies between neighboring states of their time series. On the other hand, a family of similarities based on dynamic programming and well-known to the communities of speech, bioinformatics and text-processing has been taken into account to construct kernels, namely the Dynamic-Time-Warping (DTW) score [1, 2], the Smith Waterman algorithm [3] and the edit-distance [4]. Since all these criteria do take into account the two aforementioned issues, practitioners have been tempted to use them directly with SVM implementations. However, such distances cannot be translated easily into positive definite kernels, which is an important requirement of kernel machines in the training phase. Intuitively such distances do not show favorable positive definiteness properties as they rely on the computation of an optimum rather than on the construction of a feature map, an issue that was studied in both [4, 5] and [3]. Building on these references, we propose in this work a new family of kernels between time series mostly inspired by the approach of [3]. These kernels are positive definite kernels under favorable conditions, but most importantly, they incorporate by construction more information on the compared sequences than the kernels proposed in [1, 2], while requiring exactly the same computational cost. In Section 2, we define such alignment kernels, prove their positive definiteness and show that they can be computed efficiently. We follow by presenting in Section 3 experimental results on an isolated-word recognition task using a multiclass-SVM setting, where alignment kernels need to be rescaled due to a diagonal dominance issue but still show very encouraging performances.

*corresponding author: cuturi@ism.ac.jp

# 2 Alignment Kernels

We write $\mathbb{N}$ for the set of natural positive integers, that is $\{1, 2, \ldots\}$. Let $\mathbf{x} = (x_1, \ldots, x_n)$ and $\mathbf{y} = (y_1, \ldots, y_m)$ be two finite series taking values in a state space $\mathcal{X}$, that is two elements of $\mathcal{X}^\star \overset{\text{def}}{=} \cup_{i=1}^{\infty} \mathcal{X}^i$. We define the alignment kernel in the following subsection and study its theoretical and computational properties in Section 2.2 and Section 2.3.

## 2.1 A kernel inspired by the soft-max of all alignment scores

An alignment $\pi$ of length $|\pi| = p$ between two sequences $\mathbf{x}$ and $\mathbf{y}$ is a pair of increasing p-tuples $(\pi_1, \pi_2)$ such that

$$1 = \pi_1(1) \leq \cdots \leq \pi_1(p) = n,$$
$$1 = \pi_2(1) \leq \cdots \leq \pi_2(p) = m,$$

with unitary increments and no simultaneous repetitions, that is $\forall 1 \leq i \leq p-1$,

$$\pi_1(i+1) \leq \pi_1(i)+1, \quad \pi_2(j+1) \leq \pi_2(j)+1,$$
$$(\pi_1(i+1)-\pi_1(i)) + (\pi_2(i+1)-\pi_2(i)) \geq 1.$$

We write $\mathcal{A}(\mathbf{x},\mathbf{y})$ for the set of all possible alignments between $\mathbf{x}$ and $\mathbf{y}$. Intuitively, an alignment $\pi$ between $\mathbf{x}$ and $\mathbf{y}$ describes a way to associate each element of a sequence $\mathbf{x}$ to one or possibly more elements in $\mathbf{y}$, and vice-versa. Such alignments can be conveniently represented by paths in the $n \times m$ grid displayed in Figure 1. We would like to outline now an important difference between the alignments defined here and those considered in [3, 5]. The previous definition for an alignment allows tokens to repeat themselves to handle variable-length sequences while [3] and [5] consider gaps instead, that is the insertion of a generic wildcard. Gaps make sense in biological sequence analysis, where insertions and deletions of patterns appear frequently in mutations of amino-acid sequences, as well as in spike data where time series are mostly binary, while repeated states make more sense in applications such as speech, where for instance a vowel might be slightly elongated when uttered by a new speaker. This has both theoretical and practical implications, since our algorithm and its theoretical justification are slightly different than those exposed in [3]. Following the well-known principle underlying DTW scores, the authors of [2] and [1] consider the score:

$$S(\pi) = \sum_{i=1}^{|\pi|} \varphi(x_{\pi_1(i)}, y_{\pi_2(i)}),$$

where $\varphi$ is an arbitrary conditionally positive-definite kernel[1] defined on $\mathcal{X} \times \mathcal{X}$ (such as minus the squared Euclidian distance in the case where $\mathcal{X}$ is Euclidian in [2] or directly a Gaussian kernel in [1]). Dynamic programming algorithms provide an efficient way to compute the optimal path $\pi^\star$ in terms of mean-score with respect to $\varphi$,

$$\pi^\star = \operatorname*{argmax}_{\pi \in \mathcal{A}(\mathbf{x},\mathbf{y})} \frac{1}{|\pi|} S(\pi).$$

The authors of [2] use a truly c.p.d. kernel (that is non p.d.), namely minus the Euclidian distance $\varphi(x,y) = -\|x-y\|^2$, to define then a "seemingly" p.d kernel through exponentiation:

$$\begin{aligned} k_{\mathrm{DTW}_1}(\mathbf{x},\mathbf{y}) &= e^{\frac{1}{|\pi^\star|} S(\pi^\star)} \\ &= \exp\left( - \operatorname*{argmin}_{\pi \in \mathcal{A}(\mathbf{x},\mathbf{y})} \frac{1}{|\pi|} \sum_{i=1}^{|\pi|} \|x_{\pi_1(i)} - y_{\pi_2(i)}\|^2 \right), \end{aligned}$$

while the authors of [1] use instead the Gaussian kernel for $\varphi$ and directly

[1] a symmetric function $\varphi : \mathcal{X} \times \mathcal{X} \to \mathbb{R}$ is conditionally positive-definite if for any family $\mathbf{x}_1, \ldots, \mathbf{x}_N \in \mathcal{X}$ and $c_1, \ldots, c_N \in \mathbb{R}$ such that $\sum c_i = 0$, we have that $\sum_{i,j} c_i c_j \varphi(x_i, x_j) \geq 0$

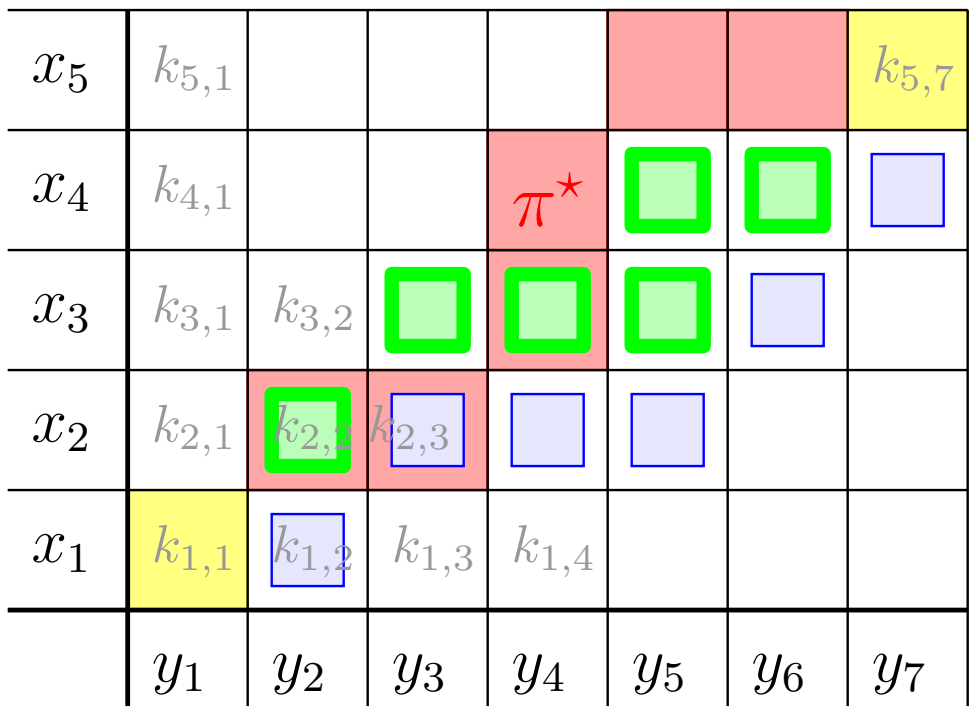

Figure 1: An alignment $\pi$ can be interpreted as a path in the $n \times m$ grid presented above, filled with the corresponding kernel values $k_{i,j} = k(x_i, y_j)$. The optimal path $\pi^\star$ is such that $\prod_{i=1}^{|\pi|} k(x_{\pi_1(i)}, y_{\pi_2(i)})$ is maximal, and corresponds in that case to $\pi_1 = (1, 2, 2, 3, 4, 5, 5, 5)$ and $\pi_2 = (1, 2, 3, 4, 4, 5, 6, 7)$. Rather than considering only the contribution of $\pi^\star$, we propose to sum up over all possible alignment paths starting from $(1, 1)$ and leading to $(5, 7)$, such as the ones represented by the two other paths.

consider[2] the corresponding score $S$ as a kernel:

$$k_{\mathrm{DTW}_2}(\mathbf{x}, \mathbf{y}) = \operatorname*{argmax}_{\pi \in \mathcal{A}(\mathbf{x},\mathbf{y})} \frac{1}{|\pi|} \sum_{i=1}^{|\pi|} e^{-\frac{1}{\sigma^2}\|x_{\pi_1(i)} - y_{\pi_2(i)}\|^2}.$$

Note that both approaches stem from a c.p.d. kernel $\varphi(x, y) = -\|x - y\|^2$ which is either exponentiated once $S$ is maximized as in [2], or directly exponentiated in the definition of $S$ to yield an optimal sum of Gaussian kernels as in [1]. In both cases, the authors aim to take advantage of such an exponentiation to turn the kernel seemingly positive definite, although this is not insured neither in theory nor in practice. We refer to the proofs of [4] and [3] to give the reader an intuition of why this is so.

The kernel we propose is not based on an optimal path chosen given a criterion $S$ induced by $\varphi$, but takes advantage of all score values $\{S(\pi), \pi \in \mathcal{A}(\mathbf{x}, \mathbf{y})\}$ spanned by all possible alignments. We argue that the following kernel is positive-definite under mild conditions and may prove more robust to quantify

[2] we have dropped to improve the readability of this presentation a few more parameters that both the authors of [1, 2] incorporate, but which do not change the overall form of the criteria they consider. We take them into account in the experimental section.

the similarity of two sequences:

$$
\begin{aligned}
K(\mathbf{x},\mathbf{y}) &= \sum_{\pi\in\mathcal{A}(\mathbf{x},\mathbf{y})} e^{S(\pi)} = \sum_{\pi\in\mathcal{A}(\mathbf{x},\mathbf{y})} e^{\sum_{i=1}^{|\pi|}\varphi(x_{\pi_1(i)},y_{\pi_2(i)})} \\
&= \sum_{\pi\in\mathcal{A}(\mathbf{x},\mathbf{y})} \prod_{i=1}^{|\pi|} k(x_{\pi_1(i)},y_{\pi_2(i)})
\end{aligned}
\tag{1}
$$

where we have written $k = e^{\varphi}$. Positive definiteness aside, the main motivation of Equation (1) is to consider the soft-max[3] of the scores of all possible alignments, rather than the simple maximum of the considered criterion. Note that the definition of the kernel $K$ does not include the log used in the definition of the softmax, but we will see in Section 3 that this logarithm is ultimately required in practice. Intuitively, the sum of Equation (1) quantifies the quality of both the optimal alignment and all the alignments which are close to it, just as the kernels presented in [6] compare two histograms through the polytope of all possible transportation plans which may map one histogram to the other, rather than considering the optimal one usually associated with the Monge-Kantorovich distance. In the sense of kernel $K$, two sequences are similar not only if they have one single alignment with high score, but do rather share a wide set of efficient alignments.

## 2.2 Positive Definiteness of the Alignment kernel

We provide in this section sufficient conditions on $k$ to prove that $K$ is a positive definite kernel.

**Theorem 1** *Let $k$ be a p.d. kernel such that $\frac{k}{1+k}$ is positive definite, then $K$ as defined in Equation (1) is positive definite.*

*Proof.* For any sequence $\mathbf{x} = (x_1,\dots,x_n) \in \mathcal{X}^\star$ and any sequence $a \in \mathbb{N}^n$ we write $\mathbf{x}_a$ for

$$\mathbf{x}_a = (\underbrace{x_1,\cdots,x_1}_{a_1 \text{ times}},\underbrace{x_2\cdots x_2}_{a_2 \text{ times}},\dots,\underbrace{x_n,\cdots,x_n}_{a_n \text{ times}}).$$

We define further for a sequence $\mathbf{x}$ of size $n$ the family $\phi_{\mathbf{s}}(\mathbf{x})$ indexed by any element $\mathbf{s} \in \mathcal{X}^\star$ as the quantity

$$\phi_{\mathbf{s}}(\mathbf{x}) = \mathrm{card}\{a \in \mathbb{N}^n : \mathbf{x}_a = s\}.$$

Note for instance that if $\mathcal{X} = \{0,1\}$, $\phi_{0001}(01) = 1$ while $\phi_{0001}(001) = 2$. Let now $\kappa$ be the following kernel on $\mathcal{X}^\star$, itself parameterized by an arbitrary p.d. kernel $\chi$ on $\mathcal{X}$ such that $|\chi| < 1$:

$$\kappa(\mathbf{x},\mathbf{y}) = \begin{cases} \prod_{i=1}^{|\mathbf{x}|}\chi(x_i,y_i) & \text{if } |\mathbf{x}| = |\mathbf{y}|, \\ 0 & \text{if } |\mathbf{x}| \neq |\mathbf{y}|. \end{cases}$$

[3]given a family of positive scalars $z = z_1, z_2, \dots, z_n$ we define the soft-max of $z$ as $\log \sum e^{z_i}$

$\kappa$ is trivially a p.d. kernel on the whole of $\mathcal{X}^\star$: simply note that for any sample $\mathbf{x}_1, \dots, \mathbf{x}_N$ of points in $\mathcal{X}^\star$, the matrix $[\kappa(\mathbf{x}_i, \mathbf{x}_j)]$ can be rearranged in a block diagonal form by sorting $\mathbf{x}_1, \dots, \mathbf{x}_N$ in increasing length, with blocks which are all positive definite. The kernel $\mathcal{K}$ defined on $\mathbf{x}, \mathbf{y} \in \mathcal{X}^\star$

$$\mathcal{K}(\mathbf{x}, \mathbf{y}) = \sum_{\mathbf{s} \in \mathcal{X}^\star} \sum_{\mathbf{s}' \in \mathcal{X}^\star} \phi_{\mathbf{s}}(\mathbf{x}) \phi_{\mathbf{s}'}(\mathbf{y}) \kappa(\mathbf{s}, \mathbf{s}')$$

is positive definite by construction, and can be rewritten as

$$\mathcal{K}(\mathbf{x}, \mathbf{y}) = \sum_{a \in \mathbb{N}^n} \sum_{b \in \mathbb{N}^m} \kappa(\mathbf{x}_a, \mathbf{x}_b),$$

where $n = |\mathbf{x}|$ and $m = |\mathbf{y}|$. We write $\epsilon$ for the sequence $1, 2, 3, ...$ and given $a \in \mathbb{N}^p$, $\epsilon_a$ for

$$\epsilon_a = (\underbrace{1, \cdots, 1}_{a_1 \text{ times}}, \underbrace{2, \cdots 2}_{a_2 \text{ times}}, \dots, \underbrace{p, \cdots, p}_{a_p \text{ times}}).$$

For two sequences of same length $u_1^p$ and $v_1^p$ we write $u \otimes v$ for $((u_1, v_1), \dots, (u_p, v_p))$. A couple $(a, b) \in \mathbb{N}^n \times \mathbb{N}^m$ defines a sequence of double indexes $\epsilon_a \otimes \epsilon_b$, which we use to express $\mathcal{K}$ as

$$\mathcal{K}(\mathbf{x}, \mathbf{y}) = \sum_{\substack{a \in \mathbb{N}^n, b \in \mathbb{N}^m \\ \|a\| = \|b\|}} \prod_{i=1}^{\|a\|} \chi((\mathbf{x}, \mathbf{y})_{\epsilon_a \otimes \epsilon_b(i)}).$$

Note now that for each couple $(a, b)$ there exists a unique alignment $\pi$ and an integral vector $v$ of adequate size such that $\pi_v = \epsilon_a \otimes \epsilon_b$ ($\pi$ is namely the sequence $\epsilon_a \otimes \epsilon_b$ stripped of all repeats, recorded in $v$), and conversely that for every couple $(\pi, v)$ there exists a unique pair $(a, b)$ such that $\pi_v = \epsilon_a \otimes \epsilon_b$. Hence, writing $\chi_{\pi(i)}$ as a short cut for $\chi(x_{\pi_1(i)}, x_{\pi_2(i)})$, we have that

$$\begin{aligned}
\mathcal{K}(\mathbf{x}, \mathbf{y}) &= \sum_{\pi \in \mathcal{A}} \sum_{v \in \mathbb{N}^{|\pi|}} \prod_{j=1}^{|\pi|} \chi((\mathbf{x}, \mathbf{y})_{\pi_v(j)}) = \sum_{\pi \in \mathcal{A}} \sum_{v \in \mathbb{N}^{|\pi|}} \prod_{j=1}^{|\pi|} \chi_{\pi(j)}^{v_j} \\
&= \sum_{\pi \in \mathcal{A}} \prod_{j=1}^{|\pi|} \left( \chi_{\pi(j)} + \chi_{\pi(j)}^2 + \chi_{\pi(j)}^3 + \cdots \right) \\
&= \sum_{\pi \in \mathcal{A}} \prod_{j=1}^{|\pi|} \frac{\chi_{\pi(j)}}{1 - \chi_{\pi(j)}}.
\end{aligned}$$

Setting now $\chi = \frac{k}{1+k}$, we recover the expression of Equation (1). ■

*Remark.* Kernels $k$ such that $\frac{k}{1+k}$ is positive definite can be trivially computed by considering first a kernel $\chi$ such that $|\chi| < 1$ and defining $k = \sum_{i=1}^{\infty} \chi^i = \chi/(1-\chi)$. If $\mathcal{X}$ is Euclidian and $\chi$ is for instance the halved Gaussian

kernel $\frac{1}{2}e^{-\frac{1}{\sigma^2}\|x-y\|^2}$, then the kernel

$$k(x,y) = \frac{\frac{1}{2}e^{-\frac{1}{\sigma^2}\|x-y\|^2}}{1-\frac{1}{2}e^{-\frac{1}{\sigma^2}\|x-y\|^2}}$$

can be directly used, and is itself numerically very similar to the Gaussian kernel. In practice, most kernels that we considered, including the Gaussian kernel and the exponential of the Gaussian kernel, have the property that $\frac{k}{1+k}$ yields positive semidefinite matrices in practice, which in an experimental context will be sufficient.

### 2.3 Computation and Factorization

We show in this section that the computation of the alignment kernel $K$ can be performed in quadratic complexity, namely in $|\mathbf{x}||\mathbf{y}|$ iterations, similarly to the naive implementation of DTW scores.

**Theorem 2** *Given* $\boldsymbol{x} = (x_1,\ldots,x_n)$ *and* $\boldsymbol{y} = (y_1,\ldots,y_m)$ *two sequences of* $\mathcal{X}^\star$*, we set the double-subscripted series* $M_{i,j}$ *as* $M_{i,0} = 0$ *for* $i = 1,\ldots,n$*,* $M_{0,j} = 0$ *for* $j = 1,\ldots,m$*, and* $M_{0,0} = 1$*. Computing recursively for* $(i,j) \in \{1,\ldots,n\}\times\{1,\ldots,m\}$ *the terms*

$$M_{i,j} = (M_{i,j-1} + M_{i-1,j-1} + M_{i-1,j})\ k(x_i,y_j),$$

*we obtain that* $K(\boldsymbol{x},\boldsymbol{y}) = M_{n,m}$

*Proof.* The result can be proved by recursion and is intuitively an equivalent of the DTW algorithm where the max-sum algebra is simply replaced by the sum-product one. ■

## 3 Experiments

The proposed kernel was tested on the English E-set of the TI46 database, which consists of 3724 spoken letters from the set {B,C,D,E,G,P,T,V,Z}. The set has a predefined division into a training set and a test-set with 1433 and 2291 utterances, respectively. From each signal we extracted a sequence of 13-dimensional feature vectors with Mel-frequency cepstral coefficients (MFCC), hence $\mathcal{X}$ is simply $\mathbb{R}^{13}$ here. The feature vectors were computed every 10 ms using a 25 ms wide Hamming window.

We compare in this section three different methods to predict a letter from a signal: first, a conventional HMM approach where we estimate the parameters of an HMM model for each letter based on the training set, and use these distributions to associate to a sequence in the test-set the letter for which it has maximum likelihood. We use a left-to-right HMM model with 6 states and 5 mixtures in each state with diagonal covariance matrices. The distributions were actually estimated using the delta and acceleration coefficients (that is on elements of $\mathbb{R}^{39}$), which are known to improve their performance.

Our second and third approaches are based on a standard one-vs-all multi-class SVM, using the spider-toolbox[4]. We use the kernel proposed in [1] and the alignment kernel with $\varphi$ defined as the Gaussian kernel. In both cases the parameter $\sigma \in \{10, 15, 20, 25, 30, 35, 40\}$ of the Gaussian kernel along with the regularization constant $C \in \{10^i, i = -2, \ldots, 6\}$ of the SVM's are first selected to obtain the best cross validation (CV) error on the training set, estimated on 4 folds with 4 repeats. Facing exactly the same problem encountered in [3], we have to address the fact that the values of the alignment kernel are exceedingly diagonally dominant, that is that the value of the kernel $k(\mathbf{x}, \mathbf{x})$ for a point against himself is many orders of magnitude larger than $k(\mathbf{x}, \mathbf{y})$. Hence, and although this operation is known not to conserve positive definiteness, we directly use the logarithm of the alignment kernel $\log K$ to rescale the obtained values. In such a case, we do exactly consider the soft-max of the set of all $S(\pi)$ values, $\pi$ spanning $\mathcal{A}(x, y)$. The empirical Gram matrix obtained on the training set was regularized by adding to it minus its smallest eigenvalue times the identity matrix to turn all its eigenvalues positive, while the train versus test-sets kernel matrix was left unchanged . The same procedure was also conducted for the DTW kernel proposed in [1] which also produces negative eigenvalues.

We obtained a test error of 11.7% for the HMM approach, 11.5% for the kernel of [1] ($\sigma = 15$, with 10.3% CV error on the training set) and 5.4% for the alignment kernel ($\sigma = 25$ and 4.3% for train CV error), giving to the log-alignment kernel a fair edge. The regularization parameter $C$ did not have a strong influence on our results when set in the middle range and was fixed at $C = 1000$. To compare further the two kernels, we performed a 4-fold cross validation with 4 repeats on the merged train and test sets, and plot in Figure 2 the CV error of each kernel as a function of $\sigma$ to illustrate the influence of the parameter on overall results. We reimplemented the CV feature of Spider to make sure for both kernels that the regularization was only carried out on the training-fold Gram matrix. Figure 2 clearly advocates the soft-max perspective provided by the log-alignment kernel.

[4]see http://www.kyb.tuebingen.mpg.de/bs/people/spider/

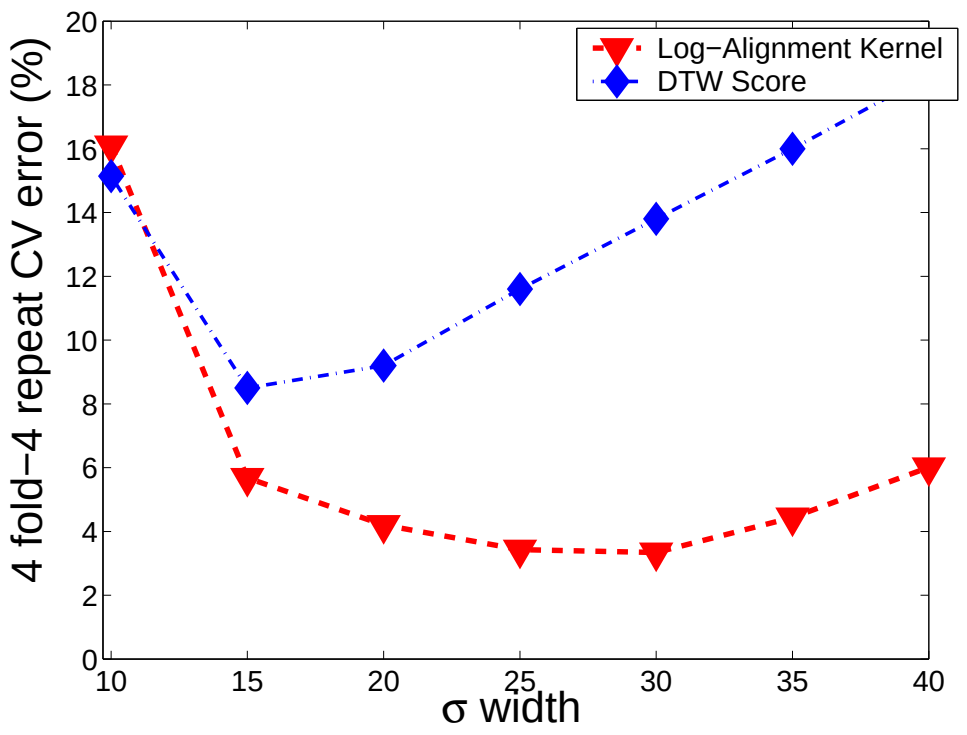

Figure 2: 4-fold with 4-repeats cross validation errors on the whole dataset of 3724 utterances as a function of the $\sigma$-Gaussian kernel width for the two studied kernels. All CV standard deviations are below a tenth of the presented values.